\documentclass[11pt]{article}

\usepackage[final]{acl}

\usepackage{times}
\usepackage{latexsym}

\usepackage[T1]{fontenc}

\usepackage[utf8]{inputenc}

\usepackage{microtype}

\usepackage{inconsolata}

\usepackage{graphicx}

\usepackage{booktabs}
\usepackage[table]{xcolor}
\usepackage{tabularx}
\usepackage{amsmath}
\usepackage{xcolor}
\usepackage{graphicx}
\usepackage{subcaption}
\usepackage{caption}
\usepackage[cjk]{kotex}
\usepackage{wasysym}
\usepackage{makecell}
\usepackage{multirow} 
\usepackage{mdframed}
\usepackage{tablefootnote}
\usepackage{longtable}
\usepackage{setspace}
\usepackage{amssymb}

\usepackage{bbding}
\usepackage{enumitem}
\usepackage[normalem]{ulem}

\usepackage{listings}
\useunder{\uline}{\ul}{}
\newcommand\ours{CORAL}

\definecolor{UpRed}{RGB}{200,0,0}
\definecolor{DownBlue}{RGB}{0,70,200}
\definecolor{SameBlack}{RGB}{0,0,0}

\mdfsetup{
linecolor=white,
backgroundcolor=gray!20,
}
\title{CORAL: Adaptive Retrieval Loop for Culturally-Aligned\\Multilingual RAG}

\author{
Nayeon Lee$^{\clubsuit}$\thanks{Equal contribution.}\thanks{Work done at Samsung Research.},
Jiwoo Song$^{\diamond*}$,
Byeongcheol Kang$^{\diamond*}$\\[0.5em]
$^{\clubsuit}$Naver, $^{\diamond}$Samsung Research\\[0.5em]
\texttt{nayeonly.lee@navercorp.com, \{jiwoooo.song, bc1.kang\}@samsung.com}
}

\begin{document}
\maketitle

\begin{abstract}

Multilingual retrieval-augmented generation (mRAG) is often implemented within a fixed retrieval space, typically via query or document translation or multilingual embedding vector representations. However, this approach may be inadequate for culturally grounded queries, in which retrieval-condition misalignment may occur. Even strong retrievers and generators may struggle to produce culturally relevant answers when sourcing evidence from inappropriate linguistic or regional contexts. To this end, we introduce \textbf{\ours{}} (\textbf{C}\textbf{O}ntext-aware \textbf{R}etrieval with \textbf{A}gentic \textbf{L}oop, an adaptive retrieval methodology for mRAG that enables iterative refinement of both the retrieval space (corpora) and the retrieval probe (query) based on the quality of the evidence. The overall process includes: (1) selecting corpora, (2) retrieving documents, (3) critiquing evidence for relevance and cultural alignment, and (4) checking sufficiency. If the retrieved documents are insufficient to answer the query correctly, the system (5) reselects corpora and rewrites the query. Across two cultural QA benchmarks, \ours{} achieves up to a 3.58\%p accuracy improvement on low-resource languages relative to the strongest baselines.
\end{abstract}
\section{Introduction}
\begin{figure}[t!]
    \centering
    \includegraphics[width=\columnwidth]{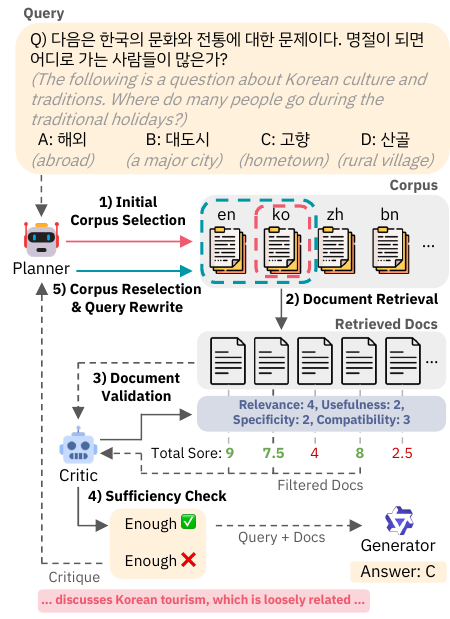}
    \caption{\textbf{Overview of \ours{}.} At test time, \ours{} performs feedback-driven retrieval control: (1) a planner selects culturally/linguistically relevant corpora, (2) retrieves top-$K$ documents, (3) a critic scores and filters them, (4) checks evidence sufficiency, and if insufficient, (5) revises corpus selection and rewrites the retrieval query based on the critique before iterating and generating.}
    \label{fig:main_figure}
\end{figure} 


Retrieval-augmented generation (RAG) improves factual grounding by incorporating external knowledge at inference time, without retraining the language model \cite{lewis2020retrieval, ovadia-etal-2024-fine}. Multilingual RAG (mRAG) extends this paradigm to support linguistically diverse queries, commonly through query translation or multilingual dense retrieval in shared embedding spaces \cite{liu-etal-2025-xrag, zhang-etal-2023-miracl, chirkova-etal-2024-retrieval}. These approaches improve linguistic coverage, but typically assume fixed retrieval conditions.


However, mRAG systems often struggle with culturally or regionally grounded queries, where factual correctness depends on local institutions, conventions, or culturally specific terminology. In such cases, systems retrieve evidence that is semantically relevant yet culturally misaligned, yielding answers that are formally correct but pragmatically inaccurate \cite{longpre-etal-2021-mkqa, li-etal-2024-bordirlines, cruz-blandon-etal-2025-memerag}. This failure mode is commonly driven by globally aggregated corpora that underrepresent locale-specific knowledge \cite{park-lee-2025-investigating, qi-etal-2025-consistency, li2025languagedrift}. As a result, errors often stem not from generation itself, but from retrieval that is misaligned with the cultural context of the query.


Existing agentic mRAG methods primarily focus on how to search---e.g., iterative query reformulation or reasoning-driven retrieval---while operating under fixed retrieval conditions \cite{asai2024selfrag, trivedi-etal-2023-interleaving, yao2023react}. As a result, query-only adaptation is often insufficient for culturally grounded queries, repeatedly sufacing culturally dominant but locale-mismatched evidence. We argue that effective multicultural mRAG requires \textbf{retrieval-condition adaptation}, where both retrieval scope and query formulation are dynamically revised based on feedback from retrieved evidence.

To this end, we propose \textbf{\ours{}} (\textbf{CO}ntext-aware \textbf{R}etrieval with \textbf{A}gentic \textbf{L}oop), a multilingual and multicultural agentic RAG framework. \ours{} iteratively adapts retrieval conditions at test time by (i) selecting query-conditioned corpora, (ii) rewriting retrieval queries via evidence critique, and (iii) explicitly checking evidence sufficiency before generation. This enables more reliable grounding for culturally specific questions. 


We evaluate \ours{} on two culturally grounded multiple-choice QA benchmarks spanning high-, mid-, and low-resource languages, covering a total of 13 languages. \ours{} consistently outperforms multilingual RAG baselines, achieving up to 3.58\%p accuracy improvement on low-resource languages relative to the strongest baselines. These results demonstrate that feedback-guided adaptation of retrieval conditions is critical for reliably grounding culturally specific answers.

Our contributions are threefold:
\begin{itemize}
\item{We identify retrieval condition misalignment as a primary failure mode of mRAG on culturally grounded queries, and reframe multilingual retrieval as feedback-driven retrieval control.}
\item{We propose \ours{}, an agentic framework that jointly adapts retrieval corpora and performs planner-guided query rewriting, with an explicit evidence sufficiency check.}
\item{We demonstrate consistent gains on culturally grounded QA benchmarks, showing that \ours{} reliably identifies target cultures across diverse languages.}
\end{itemize}

\section{Background and Related Work}
\subsection{Multilingual and Cross-lingual RAG}

Prior work on multilingual RAG mainly focuses on extending English-centric RAG pipelines to multiple languages through shared multilingual retrievers, translation-based methods, and cross-lingual benchmarks \cite{chirkova-etal-2024-retrieval, moon-etal-2025-quality, liu-etal-2025-xrag}. These approaches aim to improve linguistic coverage and robustness by preserving semantic equivalence across languages, typically operating over a fixed multilingual corpus that pools documents from all languages together.

While effective for many general cross-lingual tasks, this paradigm treats multilinguality as a representation or preprocessing problem and leaves corpus selection implicit. As a result, retrieval is performed without explicit consideration of cultural or regional relevance, which can lead to semantically relevant but culturally mismatched evidence for queries grounded in local institutions or conventions \cite{qi-etal-2025-consistency, ranaldi2025multilingual}.

\subsection{Iterative and Agentic Retrieval for RAG}

Recent work has explored iterative and agent-based retrieval strategies to improve retrieval-augmented generation 
\cite{asai2024selfrag, trivedi-etal-2023-interleaving, yao2023react, wang2024rat, yuan2024hybridrag, li-etal-2025-brief, liu2025hmrag, besrour2025ragenta}. These approaches introduce multiple retrieval steps, planning mechanisms, or specialized agents such as planners, critics, or verifiers. 
Common techniques include query reformulation, multi-hop retrieval, and retrieval planning, where the system refines its queries based on intermediate results \cite{lee-etal-2024-planrag, chen2025mao, cong-etal-2025-query}.

The main goal of these methods is to improve retrieval quality by increasing coverage, recall, or reasoning depth. 
Iteration is typically used to retrieve more relevant documents, reduce noise, or better support complex reasoning tasks \cite{asai2024selfrag, zhang2025ratt}. In this setting, agentic components help decide how to search, such as which query to issue next or when to stop retrieving \cite{yao2023react}. Some approaches further enhance these decision-making processes through additional training, such as reinforcement learning or supervised fine-tuning, enabling the model to learn when and how to retrieve more effectively \cite{asai2024selfrag, huang-etal-2025-selfaug, yao-etal-2025-seakr}. However, such approaches introduce additional training cost and may suffer from limited adaptability when the retrieval conditions themselves, such as the underlying corpus or language setting, are misaligned, as they do not explicitly reconsider or update the retrieval environment.

Despite these improvements, existing approaches usually assume that the retrieval space itself is fixed. While queries may be refined over multiple steps, the underlying corpus or knowledge source remains unchanged \cite{jang-etal-2024-itercqr, cong-etal-2025-query}. As a result, iteration focuses on improving document ranking or query formulation, rather than reconsidering whether retrieval is being performed over the most appropriate linguistic, regional, or cultural sources.

In contrast, our work treats iteration as a mechanism for correcting retrieval condition misalignment. Instead of only refining queries, the system evaluates whether the current retrieval setting is suitable and updates corpus selection decisions when necessary.

\subsection{Cultural Grounding and Context Sensitivity in RAG}
 
Prior studies have shown that retrieval and question answering systems often fail on queries that depend on cultural or regional context, producing answers that are plausible but inappropriate for the user’s setting, particularly in low-resource languages and regions \cite{li-etal-2025-multilingual, park-lee-2025-investigating, lertvittayakumjorn-etal-2025-towards}. Most existing work addresses cultural grounding through dataset construction or output analysis, while leaving the retrieval process unchanged \cite{blodgett-etal-2020-language, liu-etal-2025-xrag, thakur-etal-2025-mirage}.

While some approaches rely on query rewriting or translation \cite{chan2024rqrag, wang2025maferw}, such strategies operate within a fixed retrieval space and cannot correct cultural misalignment when relevant evidence is absent or dominated by globally prevalent sources. As a result, cultural relevance is treated as a post-hoc generation issue rather than a retrieval-time decision, even though semantically relevant documents may lack the contextual grounding required for culturally specific queries \cite{amirshahi2025evaluating, cruz-blandon-etal-2025-memerag}.


\section{Agentic Multicultural RAG}

\subsection{Overview}
We propose \ours{}, a test-time framework for culturally grounded mRAG. \ours{} comprises two LLM-based agents: a \textbf{planner} that controls corpus selection and query reformulation, and a \textbf{critic} that evaluates retrieved documents and controls sufficiency. Together, they form a feedback loop that iteratively refines retrieval conditions based on evidence quality (Figure~\ref{fig:main_figure}).

Given an input query, \ours{} executes a five-step retrieval-control loop.
\textbf{(1) Query-conditioned corpus selection:} the planner selects a small set of culturally and linguistically relevant corpora, rather than retrieving from a fixed pooled multilingual space.
\textbf{(2) Evidence retrieval:} the retriever retrieves top-$K$ documents from the selected corpora.
\textbf{(3) Critique-guided evidence validation:} the critic scores each document along multiple dimensions (relevance, usefulness, clarity/specificity, and contextual compatibility) and filters low-quality evidence.
\textbf{(4) Sufficiency checking:} the critic determines whether the retained evidence is sufficient to answer the query reliably.
\textbf{(5) Retrieval condition refinement:} if evidence is insufficient or misaligned, the critique is fed back to the planner, which revises the retrieval conditions by re-selecting corpora and reformulating the retrieval query, and repeats the loop.



\subsection{Planner: Retrieval Condition Selection and Query Reformulation}
Given the query (and, in later rounds, feedback from the critic), the planner outputs (i) a set of target corpora and (ii) an optional rewritten retrieval query (from the second round). Corpus selection is explicitly query-conditioned: the planner mainly includes corpora matching the query language and may add additional corpora when the query contains cultural or regional cues (e.g., local institutions, conventions, or region-specific entities). This scoping reduces noise from unrelated corpora and increases the likelihood of retrieving culturally grounded evidence.

When the critic indicates that retrieved evidence is insufficient or misaligned, the planner updates its decisions using the critique given. It may revise the corpus scope (e.g., expand to culturally adjacent corpora to recover missing local evidence, or narrow the scope to reduce irrelevant retrieval) or reformulate the query to address failures identified by the critique. Note that query reformulation goes beyond translation: it can clarify implicit constraints, disambiguate context-dependent terms, and introduce missing local cues surfaced during critique. This iterative planning progressively corrects retrieval condition misalignment across rounds.

\subsection{Critic: Evidence Validation and Sufficiency Control}
Following \citet{levine2025relevance}, which demonstrated that reranking documents beyond simple relevance can improve RAG systems, we introduce a multi‑dimensional scoring scheme tailored to our framework. The critic model evaluates each retrieved document and outputs (i) scores on four criteria---\textit{relevance}, \textit{usefulness}, \textit{clarity/specificity}, and \textit{contextual compatibility}---and (ii) a concise textual critique. Documents that fall below a predefined quality threshold are discarded, while those that satisfy all criteria are retained and accumulated across iterations as validated evidence for generation. Detailed definitions of the four criteria are given in Appendix~\ref{sec:scoring}, and the specifics of our scoring and filtering procedures are described in Section~\ref{sec:setup}.

After scoring, the critic determines whether the current validated evidence set is adequate to answer the query reliably. If key constraints are missing, evidence is contradictory, or alignment remains weak, the system triggers another iteration and passes the critique back to the planner. Otherwise, the loop terminates and the generator produces the final answer using only validated evidence. By coupling per-document validation with an explicit overall sufficiency decision, \ours{} performs feedback-driven retrieval control entirely at test time, without fine-tuning and with minimal assumptions about the underlying retriever or generator.

\begin{table*}[t]
\resizebox{\textwidth}{!}{
\centering
\small
\begin{tabular}{l ccccccc c}
\toprule
\multirow{3}{*}{\textbf{Method}} &
\multicolumn{7}{c}{\textbf{BLEnD}} &
\multirow{3}{*}{\textbf{CLIcK}}\\
&\multicolumn{2}{c}{\textbf{low}}&\multicolumn{2}{c}{\textbf{mid}}&\multicolumn{2}{c}{\textbf{high}}&\textbf{all}\\
\cmidrule(lr){2-8}
& su&avg & fa&avg & es&avg &avg \\
\midrule
Non-RAG & 58.04 & 55.65 & 62.09 & 63.06 & 68.59 & 69.29 & 62.13 & 48.10 \\
\addlinespace[1pt]
\cmidrule(lr){1-9}
monoRAG & 57.69 & 56.80 & 65.03 & 65.47 & 68.44 & 71.31 & 63.93 & 53.53 \\
tRAG & - & - & - & - & - & - & - & 56.06 \\
multiRAG & 61.89 & 56.48  & 67.97 & 65.92 & 67.98 & 69.84 & 63.49 & 50.78 \\
crossRAG & 62.59 & 57.83 & 67.32 & 66.83 & 68.29 & 69.76 & 64.27 & 53.75 \\
\addlinespace[1pt]
\cmidrule(lr){1-9}
\textbf{\ours{} (\textsc{gpt-oss-120b})} &
\textbf{68.18} &
\uline{60.47} &
\uline{70.92} &
\uline{69.10} &
\textbf{74.36} &
\textbf{73.51} &
\uline{67.14} & 
\uline{58.66} \\
\textbf{\ours{} (\textsc{Qwen3-235B})} &
\uline{66.78} &
\textbf{61.83} &
\textbf{72.22} &
\textbf{70.41} &
\uline{71.93} &
\uline{72.76} &
\textbf{67.84} & 
\textbf{58.88} \\
\bottomrule
\end{tabular}
}
\caption{\textbf{Accuracy on cultural QA benchmarks with \textsc{Llama-3.2-3B-Instruct}}. For \ours{}, we use \textsc{gpt-oss-120b} or \textsc{Qwen3-235B-A22B-Instruct} as the planner/critic. Best results are in \textbf{bold}, and second best results are \uline{underlined}. \ours{} improves performance by enabling dynamic corpus selection and query rewriting compared to other RAG methods that use a fixed set of target corpora.}
\label{tab:main_result_table_new}
\end{table*}
\section{Experiments}
\subsection{Datasets}

To evaluate the effectiveness of our framework, we curate multilingual QA benchmarks that require culturally grounded knowledge and commonsense reasoning without paired evidence documents.

\paragraph{BLEnD}\cite{NEURIPS2024_8eb88844} evaluates everyday cultural knowledge for 16 countries, including under-represented regions and low-resource language communities (e.g., Assam, West Java). 
We use its multiple-choice (MCQ) subset, where each question is written in English but is associated with a specific target country/culture and one of 13 source-language communities. 
Because the same underlying prompt can appear with multiple country-specific option sets, we sample one instance per underlying question to avoid overweighting duplicated prompts; full details are provided in Appendix~\ref{sec:blend_appendix}.

\paragraph{CLIcK}\cite{kim-etal-2024-click} consists of Korean MCQs gathered from official exams in addition to those generated through GPT-4~\cite{openai2024gpt4technicalreport} based on official educational materials provided by the Korean Ministry of Justice. As our focus is on cultural QAs, we use the Culture category from CLIcK. This category includes 8 subcategories including Korean Tradition, Korean Society, and Korean Popular Culture, including 1,345 queries in total. A detailed statistics of the number of questions for each subcategory can be found in Appendix~\ref{sec:click_appendix}.

\subsection{Baselines}
\label{sec:rag_methods}
To evaluate \ours{}, we compare against one non-retrieval baseline and four multilingual RAG configurations adapted from \citet{ranaldi2025multilingual}. These baselines vary the retrieval scope and translation strategy, allowing us to isolate the effects of corpus/language selection under a fixed generator.

\textbf{Non-RAG} answers the question directly without external retrieval, relying solely on the generator's internal knowledge.
\textbf{tRAG} (translate-then-retrieve) translates the query into English and retrieves only from the English corpus. As the MCQ subset of BLEnD is already in English, we present only the results on CLIcK for this baseline methodology.
\textbf{monoRAG} retrieves from the corpus that matches the query language.
\textbf{multiRAG} retrieves from the entire existing multilingual corpus without any corpus restriction.
\textbf{crossRAG} retrieves from the same corpus pool as \textbf{multiRAG}, but translates the retrieved documents into English prior to answer generation.

For query and document translation in \textbf{tRAG} and \textbf{crossRAG}, we use \textsc{Qwen3-235B-A22B-Instruct-2507}~\cite{qwen3technicalreport} (hereafter, \textsc{Qwen3-235B}).


\subsection{Experimental Setup}
\label{sec:setup}
\paragraph{Retrieval.}
For all RAG-based methods, we embed documents with \textsc{Qwen3-Embedding-8B}~\cite{qwen3embedding} and retrieve the top-$5$ documents by cosine-similarity nearest-neighbor search using FAISS~\cite{douze2024faiss}. Retrieval is performed over the target corpus scope specified by each method.

In \ours{}, the planner selects a query-conditioned set of target corpora. Then, we retrieve the top-$5$ documents from each selected corpus and pass them to the critic, which assigns integer scores in $[0,5]$ for four dimensions: relevance ($s_{rel}$), usefulness ($s_{use}$), clarity/specificity ($s_{spec}$), and contextual compatibility ($s_{comp}$). A document is considered valid if (i) each score is at least $2$ and (ii) the aggregated score $s_{tot}$ is at least $6$, where $s_{tot}$ is calculated based on the following equation:
\begin{equation}
\label{eq:total_score}
s_{tot}=s_{rel}+0.5\,(s_{use}+s_{spec}+s_{comp})
\end{equation}
Validated documents are accumulated across iterations of the feedback loop.

After the loop terminates, we select the top-$5$ validated documents by $s_{tot}$ and provide them to the generator as evidence, controlling context length while retaining the highest-quality support.

\paragraph{Inference Settings. }
In principle, any language model can serve as the planner or the critique agent. However, for our experiments, we use the same model for both planner and critique agents in our experiments. We use \textsc{Qwen3-235B}~\cite{qwen3technicalreport} and \textsc{gpt-oss-120b}~\cite{openai2025gptoss120bgptoss20bmodel} as our main planner/critic model, and \textsc{Llama-3.2-3B-Instruct}~\cite{grattafiori2024llama3herdmodels} as our main generator model.
All prompts and specific configuration details are provided in Appendix \ref{sec:prompts} and \ref{sec:model_configs}.

\paragraph{Retrieval Corpus Selection. }
Due to limited computational resources, we limit our retrieval language corpus to languages that appear as source languages in the BLEnD MCQ set.
This whole language set also covers the CLIcK dataset, which is constructed in Korean.
BLEnD is created by collecting everyday-life questions from 16 countries in 13 languages, and the MCQ subset is based on the English versions of those questions.
To ensure that every required source language is represented, we extract the Wikipedia dumps\thinspace\footnote{We use the Wikipedia dump as of October 20, 2025.} for the same 13 languages and treat them as our overall corpus. This multilingual corpus is used for the \textbf{multiRAG} and \textbf{crossRAG} approaches as the overall target corpus. The language list is provided to the planner model for query-conditioned corpus selection.


\section{Results and Analysis}

\begin{figure*}[t]
  \centering
  \includegraphics[width=\textwidth]
  {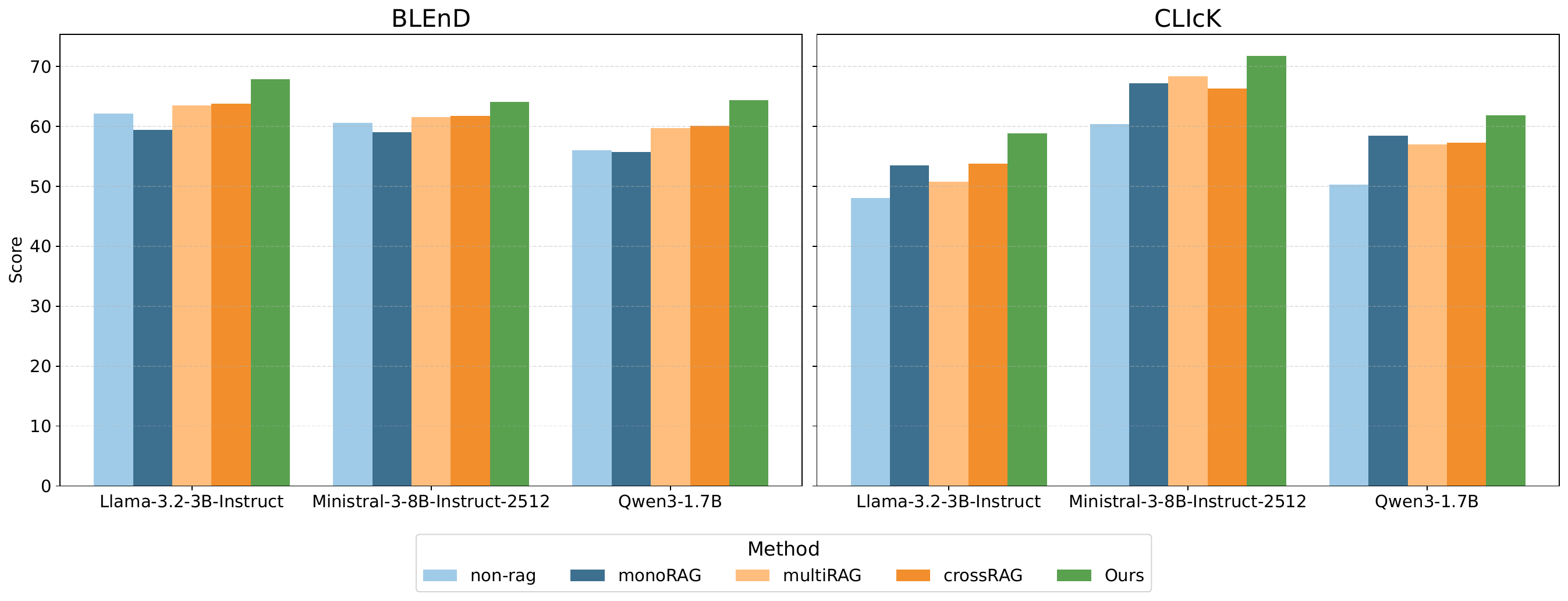}
  \caption{\textbf{Accuracy across three language models on cultural QA benchmarks.} Performance gaps between RAG baselines highlight the adverse impact of indiscriminate corpus expansion, whereas our method consistently outperforms the other baselines across diverse model families and parameter sizes.}
  \label{fig:main_result_figure}
\end{figure*}

\subsection{Overall Performance on Cultural Benchmarks}


Table~\ref{tab:main_result_table_new} reports end-to-end accuracy on two culturally grounded QA benchmarks. 
For BLEnD, we evaluate the MCQ subset in which all questions are written in English while the underlying cultural target varies across countries (Appendix~\ref{sec:blend_appendix}). 
Following the language-resource taxonomy of \citet{joshi-etal-2020-state}, we group BLEnD source-language communities into three resource tiers based on the five-level ranking: low-resource (ranks 1--2), mid-resource (ranks 3--4), and high-resource (rank 5). 
We report both the tier-wise averages and representative languages from each tier (Sundanese (su)\thinspace\footnote{A language spoken in West Java, Indonesia.}, Persian (fa), and Spanish (es) for low-, mid-, and high-resource, respectively), together with per-language results.
Results for all 13 source languages in BLEnD can be found in Appendix~\ref{sec:detailed_results}.

Across both benchmarks, \ours{} achieves the best accuracy across the two planner/critic backbones and for all resource tiers. This indicates that the gains are not tied to a specific agent model family or to a particular language group. 
To quantify improvements, we compare \ours{} against the strongest non-\ours{} baseline for each setting (i.e., the highest-scoring method among the baselines in the same column). 
On BLEnD, when using the \textsc{Qwen3-235B} planner/critic model, \ours{} gains up to 3.58\% accuracy on low-resource languages on average, especially improving the performance up to 5.59\%p for su. On CLIcK, the maximum gain reaches 3.91\%p. 
Compared with Self-RAG~\cite{asai2024selfrag}, another agentic RAG approach, our method achieves consistently better performance, with gains up to 12.14\%p. Detailed comparison is provided in Table \ref{tab:appendix_SelfRAG_comparison_table}.

Importantly, these improvements are not explained by just using more language corpora, or by a single retrieval heuristic. 
Baselines relying on a fixed retrieval scope (monoRAG/multiRAG), or with additional one-shot translation pipeline (tRAG/crossRAG) remain substantially behind, suggesting that indiscriminate corpus expansion or direct translation alone is insufficient for culturally grounded QA. 
In contrast, \ours{} couples query-conditioned corpus scoping with critique-guided query rewriting in a feedback loop. 
When the current evidence document set is incomplete or culturally misaligned, the planner revises both \emph{where} to retrieve (the corpus scope) and \emph{what} to retrieve (the retrieval query).
For example, the planner reformulates the query to match the selected corpus language, or narrows down the focus in order to retrieve a better result. 
This joint, iterative adaptation improves evidence quality and provides consistent end-to-end improvement across cultural benchmarks.



\subsection{Robustness Across Model Families and Size of the Generators}
Figure~\ref{fig:main_result_figure} shows the accuracy scores for each of the methods from Table~\ref{tab:main_result_table_new} across different generator models with varying model family and size. 
We report the average performance over all languages from BLEnD in this section.\thinspace\footnote{Figure~\ref{fig:main_result_figure} reports results for three representative generators; comprehensive evaluations across all 6 generators are provided in Appendix~\ref{sec:detailed_results}.} Across diverse model architectures and sizes, \ours{} consistently improves accuracy on the two benchmarks. 
This suggests that the observed improvements primarily originate from our feedback-driven agentic loop with minimal dependence on the generator's ability. 


\begin{figure}[t!]
  \centering
  \includegraphics[width=\columnwidth]
  {Figures/language_distribution.pdf}
  \caption{\textbf{Language distribution of documents selected for RAG.} Hatched bars indicate the language proportions of documents selected by the planner for each benchmark, while solid bars represent the language distribution of documents actually used for RAG after critique-based scoring.}
  \label{fig:language_distribution}
\end{figure}

\subsection{Analysis on Dynamic Corpus Selection}
\label{sec:dynamic_corpus_selection}

\begin{table}[t]
\resizebox{\columnwidth}{!}{
\centering
\setlength{\tabcolsep}{6pt}
\begin{tabular}{l ccccccc c}
\toprule
\multirow{2}{*}{\textbf{Method}} &
\multicolumn{3}{c}{\textbf{BLEnD}} &
\multirow{2}{*}{\textbf{CLIcK}}\\
\cmidrule(lr){2-4}
&\textbf{low}&\textbf{mid}&\textbf{high}&\\
\midrule
Non-RAG & 55.65 & 63.06 & 69.29 & 48.10 \\
RAG$_{\mathcal{C}_{own}}$      & 51.89 & 60.77 & 67.43 & 53.53\\
RAG$_{\mathcal{C}_{all}}$      & 56.55 & 65.92 & 69.84 & 50.78 \\
RAG$_{\mathcal{C}_{own}\,\cup\,\mathcal{C}_{en}}$   & 56.06 & 65.94 & 71.22 & 54.20 \\
\midrule
\textbf{\ours{}} & \textbf{61.83} & \textbf{70.41} & \textbf{72.78} & \textbf{58.88} \\
\bottomrule
\end{tabular}
}
\caption{\textbf{Static corpus ablation on cultural QA benchmarks.} 
We compare fixed retrieval scopes: $\mathcal{C}_{own}$ (the culture-associated corpus; for BLEnD we use the source language), $\mathcal{C}_{all}$ (overall multilingual corpora), and $\mathcal{C}_{own}\cup\mathcal{C}_{en}$ (adding English). 
Fixed corpus scopes show inconsistent gains across benchmarks and resource groups, while \ours{} (\textsc{gpt-oss-120b} planner/critic) remains consistently stronger.}

\label{tab:corpus_ablation}
\end{table}

\begin{table*}[t]
\centering
\small
\setlength{\tabcolsep}{15pt}
\begin{tabular}{l ccc c}
\toprule
\multirow{2}{*}{\textbf{Method}} &
\multicolumn{3}{c}{\textbf{BLEnD}} &
\multirow{2}{*}{\textbf{CLIcK}}\\
\cmidrule(lr){2-4}
&\textbf{low}&\textbf{mid}&\textbf{high}&\\
\midrule
multiRAG  & 56.55 & 65.92 & 69.84 & 50.78  \\ \midrule
\multicolumn{5}{l}{w/ \textsc{gpt-oss-120b} Planner/Critic} \\ \midrule
\textit{+ Dynamic Corpus Selection} & 58.11 & \textbf{70.06} & 72.76 & 57.25 \\
\hspace{3mm}\textit{+ Query Rewriting} (\ours{}) & \textbf{60.47} & 69.10 & \textbf{73.51} & \textbf{58.66} \\ \midrule
\multicolumn{5}{l}{w/ \textsc{Qwen3-235B} Planner/Critic}   \\ \midrule
\textit{+ Dynamic Corpus Selection} & 59.64 & 69.70 & 71.64 & 57.40 \\
\hspace{3mm}\textit{+ Query Rewriting} (\ours{}) & \textbf{61.83} & \textbf{70.41} & \textbf{72.78} & \textbf{58.88} \\ \bottomrule
\end{tabular}
\caption{\textbf{Ablation of dynamic corpus selection and query rewriting.} Accuracy on five cultural QA benchmarks with a fixed generator. Starting from multiRAG (a fixed-pooled multilingual retrieval system with the original query), we add dynamic corpus selection and then query rewriting. Results are shown for two planner/critic backbones.}

\label{tab:query_ablation}

\end{table*}

Figure~\ref{fig:language_distribution} visualizes the language compositions of (i) the planner-selected corpus set and (ii) the final top-$K$ evidence after critique-guided scoring, for both datasets. A key observation is that the planner’s choices go beyond query-language detection. This is most evident on BLEnD, where all queries are written in English. The planner selects the culture-associated languages and their regional high-resource neighbors, along with English. For instance, in BLEnD-su it frequently selects Sundanese (su) together with Indonesian (id), and in BLEnD-fa it additionally considers Arabic (ar). This indicates that it infers the likely cultural target from the query content and routes retrieval accordingly.

The two language distributions from (i) and (ii) are broadly consistent but not identical, reflecting the role of critique-based filtering. After scoring, the retained top-$K$ evidence shifts toward documents that actually contain useful evidence and away from weakly related documents. In low-resource settings, this can increase the share of a regional high-resource language when the targeted corpus is sparse (e.g., more \textit{id} for BLEnD-su), while still maintaining culturally aligned sources. On CLIcK, the final evidence document set remains dominated by Korean (ko), with additional support from English (en) and nearby languages. Overall, Figure~\ref{fig:language_distribution} suggests that our planner-critic loop proposes a culturally plausible candidate pool and then enforces evidence quality and cultural alignment through critique-guided filtering.

\paragraph{Fixed-Scope Retrieval Ablation. }
Figure~\ref{fig:language_distribution} shows that English often appears alongside culturally aligned corpora, which motivates a natural baseline: \textit{can the planner-critique loop simply be replaced with a fixed retrieval scope that always includes English?} Table~\ref{tab:corpus_ablation} evaluates this hypothesis by comparing three fixed-scope variants: retrieving only from an oracle own-corpus ($\mathcal{C}_{own}$), retrieving from the union of all corpora ($\mathcal{C}_{all}$), and retrieving from $\mathcal{C}_{own}\cup\mathcal{C}_{en}$, where $\mathcal{C}_{en}$ denotes the English corpus. For CLIcK, $\mathcal{C}_{own}$ corresponds to Korean. For BLEnD, while the questions are written in English, $\mathcal{C}_{own}$ is defined as the source community/culture language associated with each evaluation split (e.g., \textit{su} for BLEnD-su). We emphasize that this BLEnD definition is oracle: it presumes access to a target-culture label that is not provided at test time in realistic deployments.

Table~\ref{tab:corpus_ablation} shows that fixed-scope retrieval remains consistently below \ours{}, even when granted oracle access to $\mathcal{C}_{own}$. Adding English to $\mathcal{C}_{own}$ is not uniformly sufficient across BLEnD resource groups, and pooling all corpora ($\mathcal{C}_{all}$) can saturate when culturally or content-wise mismatched documents are included.\footnote{Figure~\ref{fig:corpus-noisy-example} illustrates a representative failure mode of $\mathcal{C}_{all}$, where retrieval returns superficially related but not decision-critical evidence. Additional qualitative examples are provided in Appendix~\ref{sec:noise_appendix}.} Moreover, on BLEnD, even the oracle-fixed $\mathcal{C}_{own}$ setting can underperform Non-RAG, consistent with the fact that culturally grounded QA often relies on proxy evidence and that sparse or weak retrieval can introduce misleading context. In contrast, \ours{} consistently outperforms all fixed-scope variants, indicating that the gains are not explained by simply including English, but by query-conditioned scope decisions coupled with feedback-driven filtering (and, as shown later, critique-guided query rewriting).

\subsection{Dynamic Corpus Selection \& Query Rewriting Ablation Study}
\paragraph{Dynamic Corpus Selection Only. }To quantify the contributions of the two key components of \ours{}---dynamic corpus selection and critique-guided query rewriting---we report the ablation results in Table~\ref{tab:query_ablation}. We use multiRAG as the baseline, which retrieves with the original query from a fixed pooled multilingual corpus, and then progressively add (i) dynamic corpus selection and (ii) query rewriting.

The results show that adding dynamic corpus selection alone improves accuracy on all benchmarks for both planner/critic backbones. With \textsc{Qwen3-235B}, dynamic selection yields gains of 5.78\%p (BLEnD-mid) and 3.21\%p (CLIcK) over \textsc{multiRAG}. These results support our claim that selecting culturally appropriate retrieval conditions substantially reduces noise from mismatched corpora and improves evidence alignment.

\paragraph{Additional Query Rewriting.}
On top of dynamic corpus selection, enabling query rewriting further improves performance. With the \textsc{gpt-oss-120b} planner/critic, query rewriting achieves additional gains of 2.36\%p on BLEnD-low and 2.21\%p on CLIcK.

To better understand the contribution of query rewriting, we analyze how the planner modifies the retrieval query during rewriting. We categorize each rewrite into one of three types: (i) \textbf{Paraphrase}, which reformulates the query into a more retrieval-friendly wording while preserving its intent; (ii) \textbf{Narrow}, which adds constraints or disambiguating details to focus retrieval; and (iii) \textbf{Expand}, which broadens the query to retrieve additional evidence when the current retrieval is judged insufficient.

We randomly sample 100 questions from CLIcK and collect all rewritten retrieval queries produced across planner-critic iterations, resulting in 158 rewritten queries. After a norming session to align category definitions, two authors independently annotate all rewrites. The initial inter-annotator agreement is Cohen's $\kappa=0.624$. Remaining disagreements are then resolved through discussion, and final labels are determined by unanimous agreement.

Overall, 53.8\% of rewrites narrow the query, and 32.9\% paraphrase it.\thinspace\footnote{An example of query rewrite within a planner-critic loop is provided in Figure~\ref{fig:query-rewrite-example}. Additional qualitative analysis on query rewriting can be found in Appendix~\ref{sec:query_appendix}.} Qualitative analysis reveals that narrowing rewrites often introduce missing contextual cues, as highlighted by the critic, when the initially retrieved documents are topically related but insufficiently informative to answer the question. This leads to subsequent retrievals that are more directly aligned with the query's informational needs. Taken together, these results suggest that query rewriting complements dynamic corpus selection by systematically improving retrieval quality through critique-guided refinement.

\section{Conclusion}

We introduce \ours{}, a test-time agentic framework that closes the loop between retrieval outcomes and retrieval decisions. \ours{} iteratively (i) selects culturally and linguistically appropriate corpora, (ii) retrieves candidate evidence, (iii) critiques documents for relevance and cultural alignment, and (iv) checks sufficiency to decide whether to stop or to refine retrieval conditions by re-selecting corpora and rewriting the query. Across five culturally grounded QA benchmarks spanning high- and low-resource languages, \ours{} consistently outperformed strong multilingual RAG baselines, with the largest improvements appearing in low-resource settings where indiscriminate corpus expansion tends to introduce noise or amplify generalized evidence.

Our findings suggest that scaling multilingual coverage alone is insufficient for culturally grounded generation, and that robust multilingual RAG systems should treat corpus scope and query formulation as first-class, revisable decisions rather than fixed configuration choices. More broadly, the retrieval condition selection viewpoint provides a principled way to integrate cultural and regional constraints into retrieval-augmented generation, complementary to advances in multilingual representations and agentic reasoning.

\section*{Limitations}

While \ours{} consistently improves performance and supports culturally grounded retrieval control, it has several limitations. First, some benchmark questions may require knowledge that is sparse or entirely absent from Wikipedia-based corpora. In such cases, retrieval failures are unavoidable regardless of the control strategy. More broadly, culturally relevant information is often procedural, experiential, or locally disseminated (e.g., informal norms or recent policy details), and may be underrepresented in encyclopedic resources.

Moreover, our corpora are restricted to language-specific Wikipedia subsets. This choice improves reproducibility, but it limits domain diversity and may bias retrieval toward perspectives that are well covered in the selected languages. Extending the corpus collection and retrieval framework to heterogeneous web-scale sources (e.g., official portals, local news, and community resources) would better reflect real-world cultural information needs.

Our evaluation focuses on multiple-choice question answering to enable controlled comparisons in the study of dynamic corpus selection and query rewriting. This setting may not capture additional failure modes that arise in open-ended or interactive scenarios, such as partially correct responses, culturally inappropriate framing, or user-dependent ambiguity. Evaluating \ours{} in open-ended generation and multi-turn information-seeking settings is an important direction for future work.

\section*{Ethical Considerations}

Our approach operates during the test phase by using retrieved documents and does not require collecting user-level data or fine-tuning models. However, when deployed in real-world retrieval contexts, systems may inadvertently access or disclose personal or sensitive information contained in documents. It is imperative that deployments comply with applicable privacy regulations, implement access controls, refrain from retrieving private data without proper authorization, and accommodate data deletion requests when appropriate.

We acknowledge that agentic retrieval incurs additional inference costs due to the requirements of iterative planning and critique. While we restrict the number of iterations and permit early termination when sufficient evidence is available, practitioners should carefully weigh the efficiency trade-offs and carbon footprint associated with these processes. Future research should investigate the potential for lightweight critics, caching mechanisms, and cost-aware stopping policies to mitigate computational overhead.


\bibliography{custom,anthology-1,anthology-2}

\appendix

\section*{Appendix}
\label{sec:appendix}
\section{Critic Scoring Criteria Definitions}
\label{sec:scoring}
For each criterion, the critic assigns an integer score from 0 to 5. We define each criterion as follows.

\paragraph{Relevance.}
Relevance measures how strongly a document aligns with the key concepts, entities, and intent of the query. Higher scores indicate close conceptual alignment and direct topical relevance, while lower scores reflect only weak or incidental connections to the query.

\paragraph{Usefulness.}
Usefulness measures how much a document helps the system construct a correct, complete, and actionable answer. Higher scores indicate that the document contributes substantial, high-impact information needed for solving the query, whereas lower scores indicate little to no helpful content for answering.

\paragraph{Clarity and Specificity.}
Clarity and Specificity measures how clearly, precisely, and unambiguously a document presents information that is relevant to the query. Higher scores correspond to well-structured, specific, and easy-to-interpret statements, while lower scores correspond to content that is vague, overly general, or difficult to apply.

\paragraph{Compatibility.}
Compatibility measures linguistic, cultural, and domain alignment between the query and the document. Higher scores indicate strong language match or faithful cross-lingual equivalence, along with contextual appropriateness for the query’s cultural and domain assumptions. Lower scores indicate mismatched language, cultural context, or domain framing that makes the evidence less applicable.

\section{Detailed Description of the Datasets}  

\subsection{BLEnD}
\label{sec:blend_appendix}
In this paper, we use a subset of the multiple‑choice‑question (MCQ) data provided by BLEnD.    
The MCQ portion of BLEnD contains every possible option combination for each question across all countries, which leads to varying numbers of items for the same underlying question.    
Because we aim for a fair comparison and have limited resources, we randomly select a single version of each question (i.e., one country‑specific option set per question).    
Table~\ref{tab:blend_mcq_stats} summarizes the statistics of the selected MCQs.    
\begin{table}[th]
\resizebox{\columnwidth}{!}{
\centering
\setlength{\tabcolsep}{6pt}
\begin{tabular}{llc}
\toprule
\textbf{Source Lang.}         & \textbf{Country} & \textbf{\# of MCQs} \\ \midrule
\multirow{2}{*}{English (en)} & United States    & 310                 \\
                              & United Kingdom   & 304                 \\
\multirow{2}{*}{Spanish (es)} & Spain            & 325                 \\
                              & Mexico           & 334                 \\
\multirow{2}{*}{Korean (ko)}  & South Korea      & 366                 \\
                              & North Korea      & 290                 \\
Indonesian (id)               & Indonesia        & 334                 \\
Chinese (zh)                  & China            & 335                 \\
Arabic (ar)                   & Algeria          & 304                 \\
Greek (el)                    & Greece           & 320                 \\
Persian (fa)                  & Iran             & 306                 \\
Azerbaijani (az)              & Azerbaijan       & 325                 \\
Sundanese (su)                & West Java        & 286                 \\
Assamese (as)                 & Assam            & 358                 \\
Hausa (ha)                    & Northern Nigeria & 249                 \\
Amharic (am)                  & Ethiopia         & 335                 \\ \midrule
\textbf{Total}                &                  & 5,081               \\ \bottomrule
\end{tabular}
}
\caption{  
\textbf{Number of MCQs per country and source language.}   
For countries that share the same source language(en, es, ko), the MCQs are combined and reported as a single aggregated result elsewhere in the paper.  
}  
\label{tab:blend_mcq_stats}
\end{table} 

\subsection{CLIcK}
\label{sec:click_appendix}
In order to focus on cultural queries, we only use the Culture category from CLIcK.
Table~\ref{tab:click_mcq_stats} shows the statistics of the number of MCQs within each of the subcategories.
\begin{table}[th]
\centering
\setlength{\tabcolsep}{6pt}
\begin{tabular}{lc}
\toprule
\textbf{Category} & \textbf{\# of MCQs} \\ \midrule
Society           & 309                 \\
Tradition         & 222                 \\
History           & 280                 \\
Law               & 219                 \\
Politics          & 84                  \\
Economy           & 59                  \\
Geography         & 131                 \\
Pop culture       & 41                  \\ \midrule
\textbf{Total}    & 1,345               \\ \bottomrule
\end{tabular}
\caption{  
\textbf{Number of MCQs from CLIcK within the Culture Category.}   
}  
\label{tab:click_mcq_stats}
\end{table}

\section{Detailed Experimental Settings}

\subsection{Prompts}
\label{sec:prompts}

\subsubsection{Multiple Choice Question Generator Prompt}
\begin{mdframed}
Answer the following multiple choice question as clearly as possible, using the provided **Reference Evidence**. The last line of your response should be in the following format: `Answer: A/B/C/D/E' (e.g. `Answer: A').\\
\\
\# Reference Evidence\\
\{Docs\}\\
\\
\# Question\\
\{Query\}\\
\end{mdframed}

\subsubsection{Short Answer Question Generator Prompt}
\begin{mdframed}
Answer the following short answer question as clearly as possible, using the provided **Reference Evidence**. The last line of your response should be in the following format: `Answer: $[$YOUR ANSWER HERE$]$' (e.g. `Answer: cat').\\
\\
\# Reference Evidence\\
\{Docs\}\\
\\
\# Question\\
\{Query\}\\
\end{mdframed}

\subsubsection{Planner Prompt}
Figures \ref{fig:planner-prompt-template} and \ref{fig:planner-prompt-template-w-critique} present the prompts used for the planner. Figure \ref{fig:planner-prompt-template} is used to perform corpus selection upon receiving the initial user query. Figure \ref{fig:planner-prompt-template-w-critique} is used when the critique module determines that the retrieved documents are insufficient: the planner performs corpus selection and query reformulation for the next retrieval step.

\subsubsection{Critique Prompt}
Figure \ref{fig:critique-prompt-template-scoring} and \ref{fig:critique-prompt-template-sufficiency} present the prompts used for the critique. Figure \ref{fig:critique-prompt-template-scoring} is used to evaluate the retrieved documents against predefined criteria. Figure \ref{fig:critique-prompt-template-sufficiency} is used to determine whether the retrieval evidence is insufficient; based on this decision, the retrieval process is set to proceed iteratively.

\subsection{Model Configurations}
\label{sec:model_configs}
For planner and critique models, we set the temperature as 0.6, with reasoning effort `high' for the \textsc{gpt-oss} models and enable thinking for the hybrid \textsc{Qwen} models.
For the generator models, we set the temperature as 0 and top\_p as 1, with reasoning effort `low' for the \textsc{gpt-oss} models and disable thinking for the hybrid \textsc{Qwen} models.
We set the max token of each of the planner/critic models to 32768, with dynamic adaptation of the max token value if needed.
We set the max token of the generator models to 4096.


\section{Detailed Results}
\label{sec:detailed_results}

\subsection{Experimental Results in Detail}

We evaluate all configurations using 13 open and instruction-tuned LLMs, ranging from small to large language models: Qwen3-\{1.7B, 8B\}~\cite{qwen3technicalreport}, LLaMA-3.2-\{1B, 3B\}-Instruct~\cite{grattafiori2024llama3herdmodels}, Ministral-3-\{8B, 14B\}-Instruct-2512\thinspace\footnote{https://huggingface.co/collections/mistralai/ministral-3}. This diverse model suite enables robust comparison across a wide range of capacity and instruction tuning settings.
As shown in Table \ref{tab:appendix_benchmark_table_2}-\ref{tab:appendix_qwen-8B_table}, our method yields consistently strong and robust performance across a wide range of languages, covering diverse model sizes and model families.

We also compare \ours{} with Self-RAG~\cite{asai2024selfrag} in Table~\ref{tab:appendix_SelfRAG_comparison_table} to evaluate another agentic RAG method on BLEnD. We do not include CLIcK for comparison, as the input query required by CLIcK exceeds Self-RAG's maximum token limit. The results show that \ours{} outperforms Self-RAG by a significant margin of up to 12.14\%p.

\subsection{Planner Critique Examples for Document Selection and Query Rewriting}
\subsubsection{Noise Comparison between Global and Locale-Specific Corpora}
\label{sec:noise_appendix}

We present a qualitative comparison illustrating the difference in evidence quality when retrieval is performed over a global corpus versus a locale-specific corpus.
Shown in Figure \ref{fig:corpus-noisy-example}, the given query concerns a culturally grounded practice in Korea, specifically the ritual behavior performed during ancestral rites (jesa), where participants bow twice to honor their ancestors. Comparing the documents retrieved by the $\mathcal{C}_{all}$ and \ours{} reveals a clear qualitative difference in evidence relevance. Retrieval over the unified corpus yields mostly superficial or tangential information: some documents mention jesa only at a high level without describing the ritual procedure, while others are entirely unrelated despite sharing cultural keywords, covering topics such as first-birthday celebrations (doljanchi), Confucianism in general, or Chuseok rituals. In contrast, our agent successfully identifies a Korean-language document that explicitly explains the procedural steps of jesa, including the correct bowing practice. This example illustrates how indiscriminate corpus expansion introduces substantial noise for culturally specific queries, whereas our method effectively routes retrieval toward linguistically and culturally aligned sources, enabling the model to access precise procedural knowledge that is essential for answering the question correctly.


\subsubsection{Query Rewriting for Improved Evidence Relevance}
\label{sec:query_appendix}

We further provide a qualitative example illustrating how the planner-guided process improves evidence relevance across retrieval trials. The critic evaluates the initially retrieved documents—identifying those that are semantically related but lack sufficient grounding in the target context—while the planner utilizes these insights to rewrite the query, incorporating the missing contextual signals. Full details are shown in Figure \ref{fig:query-rewrite-example}.

\subsection{Efficiency and Token Cost Analysis}
\label{sec:efficiency_analysis}

We analyze the computational overhead of \ours{} in terms of iteration count and token usage. 
Although \ours{} employs an iterative planner--critic loop, the number of iterations is bounded and further reduced by a sufficiency-based early stopping mechanism.

Empirically, \ours{} converges in a small number of iterations. The average number of planner--critic iterations is 1.34 on BLEnD and 1.52 on CLIcK. Moreover, the final retrieval evidence is typically selected early in the process, at 1.56 and 1.82 iterations on average for BLEnD and CLIcK, respectively.

In terms of token consumption, the average agent-side token usage per instance is 1,807 tokens on BLEnD and 21,548 tokens on CLIcK. These results provide a quantitative characterization of the computational overhead of \ours{}, alongside the iteration statistics reported above.

\begin{table*}[t]
\centering
\small
\setlength{\tabcolsep}{12pt}
\begin{tabular}{l cc cc cc}
\toprule
\multirow{2}{*}{\textbf{Method}} &
\multicolumn{2}{c}{\textbf{Llama-3.2}} &
\multicolumn{2}{c}{\textbf{Ministral-3}} &
\multicolumn{2}{c}{\textbf{Qwen-3}} \\
\cmidrule(lr){2-3}
\cmidrule(lr){4-5}
\cmidrule(lr){6-7}
& 1B & 3B & 8B & 14B & 1.7B & 8B \\

\midrule
Non-RAG & 53.00 & 62.13 & 60.54 & 64.84 & 55.99 & 66.10 \\
monoRAG & 56.83 & 63.93 & 61.77 & 64.43 & 59.45 & 64.16 \\
multiRAG & 57.24 & 63.52 & 61.56 & 64.33 & 59.75 & 65.43 \\
crossRAG & 58.89 & 63.83 & 61.79 & 64.44 & 60.06 & 65.55 \\
\ours{} (GPT-OSS-120B) & 61.08 & 67.14 & 66.18 & 68.20 & 64.56 & 68.42 \\
\ours{} (Qwen3-235B) & 60.12 & 67.84 & 64.09 & 66.22 & 64.40 & 68.59 \\
\bottomrule
\end{tabular}
\caption{\textbf{Average Accuracy on BLEnD with various generators}. }
\label{tab:appendix_benchmark_table_2}
\end{table*}

\begin{table*}[t]
\resizebox{\textwidth}{!}{
\centering
\begin{tabular}{l ccccccccccccc c}
\toprule
\multirow{2}{*}{\textbf{Method}} &
\multicolumn{13}{c}{\textbf{BLEnD}} &
\multirow{2}{*}{\textbf{CLIcK}}\\
\cmidrule(lr){2-14}
& am & ar & as & az & el & en & es & fa & ha & id & ko & su & zh & \\

\midrule
Non-RAG  & 39.40 & 49.01 & 45.25 & 52.92 & 60.00 & 71.17 & 60.55 & 53.92 & 48.59 & 52.10 & 52.59 & 47.90 & 55.52 & 34.05 \\
monoRAG  & 46.27 & 60.20 & 52.23 & 55.38 & 59.06 & 66.78 & 63.28 & 59.80 & 47.39 & 57.49 & 56.40 & 52.45 & 62.09 & 52.12 \\
tRAG     & --    & --    & --    & --    & --    & --    & --    & --    & --    & --    & --    & --    & --    & 52.12 \\
multiRAG & 45.67 & 57.89 & 51.96 & 57.54 & 59.06 & 67.26 & 62.67 & 59.48 & 50.20 & 61.08 & 57.16 & 53.85 & 60.30 & 41.56 \\
crossRAG & 43.88 & 61.84 & 53.35 & 56.92 & 62.50 & 66.94 & 63.88 & 60.46 & 48.59 & 62.28 & 58.38 & 56.99 & 69.55 & 44.16 \\
\ours{} (GPT-OSS-120B)  & 46.27 & 60.86 & 49.72 & 55.08 & 62.50 & 73.94 & 68.89 & 68.95 & 49.80 & 66.17 & 61.13 & 61.19 & 69.55 & 48.25 \\
\ours{} (Qwen3-235B) & 47.16 & 60.20 & 48.88 & 56.00 & 62.50 & 69.22 & 65.40 & 66.67 & 49.80 & 68.26 & 62.65 & 57.69 & 67.16 & 47.29 \\
\bottomrule
\end{tabular}
}
\caption{\textbf{Accuracy on cultural QA benchmarks with Llama-3.2-1B for a generator}. }
\label{tab:appendix_llama-1B_table}
\end{table*}

\begin{table*}[t]
\resizebox{\textwidth}{!}{
\centering
\begin{tabular}{l ccccccccccccc c}
\toprule
\multirow{2}{*}{\textbf{Method}} &
\multicolumn{13}{c}{\textbf{BLEnD}} &
\multirow{2}{*}{\textbf{CLIcK}}\\
\cmidrule(lr){2-14}
& am & ar & as & az & el & en & es & fa & ha & id & ko & su & zh & \\

\midrule
Non-RAG  & 46.87 & 59.54 & 58.66 & 63.69 & 64.69 & 79.48 & 68.59 & 62.09 & 51.00 & 66.77 & 58.69 & 58.04 & 69.55 & 48.10 \\
monoRAG  & 53.73 & 67.76 & 56.15 & 65.85 & 68.44 & 77.69 & 68.44 & 65.03 & 50.60 & 64.07 & 64.33 & 57.69 & 71.34 & 56.06 \\
tRAG     & --    & --    & --    & --    & --    & --    & --    & --    & --    & --    & --    & --    & --    & 56.06 \\
multiRAG & 52.54 & 63.82 & 55.31 & 62.46 & 66.56 & 77.69 & 67.98 & 67.97 & 50.20 & 65.57 & 63.57 & 62.24 & 69.85 & 50.78 \\
crossRAG & 51.64 & 63.82 & 56.15 & 67.38 & 68.75 & 77.36 & 68.29 & 67.32 & 51.41 & 66.77 & 64.48 & 62.59 & 0.00  & 53.75 \\
\ours{} (GPT-OSS-120B)  & 53.73 & 64.47 & 56.70 & 69.54 & 68.75 & 79.97 & 74.36 & 70.92 & 54.22 & 67.66 & 69.05 & 68.18 & 75.22 & 58.66 \\
\ours{} (Qwen3-235B) & 55.82 & 66.78 & 60.34 & 68.00 & 69.69 & 78.66 & 71.93 & 72.22 & 58.23 & 70.96 & 68.75 & 66.78 & 73.73 & 58.88 \\

\bottomrule
\end{tabular}
}
\caption{\textbf{Accuracy on cultural QA benchmarks with Llama-3.2-3B for a generator}. }
\label{tab:appendix_llama-3B_table}
\end{table*}

\begin{table*}[t]
\resizebox{\textwidth}{!}{
\centering
\begin{tabular}{l ccccccccccccc c}
\toprule
\multirow{2}{*}{\textbf{Method}} &
\multicolumn{13}{c}{\textbf{BLEnD}} &
\multirow{2}{*}{\textbf{CLIcK}}\\
\cmidrule(lr){2-14}
& am & ar & as & az & el & en & es & fa & ha & id & ko & su & zh & \\

\midrule
Non-RAG  & 50.45 & 58.22 & 50.00 & 60.92 & 61.88 & 77.85 & 68.44 & 62.09 & 49.80 & 65.87 & 59.91 & 54.20 & 67.46 & 60.37 \\
monoRAG  & 48.06 & 66.12 & 53.63 & 64.31 & 66.25 & 74.76 & 62.37 & 67.97 & 47.79 & 63.17 & 63.57 & 56.99 & 68.06 & 61.41 \\
tRAG     & --    & --    & --    & --    & --    & --    & --    & --    & --    & --    & --    & --    & --    & 61.41 \\
multiRAG & 49.85 & 61.51 & 51.68 & 65.23 & 64.38 & 75.41 & 66.62 & 68.30 & 44.58 & 66.17 & 63.72 & 56.29 & 66.57 & 68.40 \\
crossRAG & 48.66 & 64.14 & 51.68 & 61.85 & 64.38 & 76.06 & 67.07 & 66.34 & 45.78 & 65.87 & 62.96 & 60.14 & 68.36 & 66.32 \\
\ours{} (GPT-OSS-120B)  & 55.82 & 69.74 & 55.31 & 66.77 & 66.88 & 78.66 & 72.84 & 72.88 & 51.41 & 69.16 & 66.01 & 63.29 & 71.64 & 72.42 \\
\ours{} (Qwen3-235B) & 57.01 & 67.76 & 51.96 & 62.46 & 65.31 & 76.22 & 68.59 & 66.01 & 46.59 & 70.96 & 67.07 & 60.14 & 73.13 & 71.75 \\
\bottomrule
\end{tabular}
}
\caption{\textbf{Accuracy on cultural QA benchmarks with Ministral-3-8B-Instruct-2512 for a generator}. }
\label{tab:appendix_ministral-8B_table}
\end{table*}

\begin{table*}[t]
\resizebox{\textwidth}{!}{
\centering
\begin{tabular}{l ccccccccccccc c}
\toprule
\multirow{2}{*}{\textbf{Method}} &
\multicolumn{13}{c}{\textbf{BLEnD}} &
\multirow{2}{*}{\textbf{CLIcK}}\\
\cmidrule(lr){2-14}
& am & ar & as & az & el & en & es & fa & ha & id & ko & su & zh & \\

\midrule
Non-RAG  & 53.13 & 61.84 & 57.82 & 67.08 & 64.69 & 81.92 & 68.44 & 66.01 & 56.63 & 64.97 & 63.57 & 60.14 & 76.72 & 64.31 \\
monoRAG  & 54.33 & 62.83 & 56.98 & 67.38 & 67.81 & 78.01 & 67.37 & 67.97 & 51.81 & 68.26 & 65.70 & 58.39 & 70.75 & 63.20 \\
tRAG     & --    & --    & --    & --    & --    & --    & --    & --    & --    & --    & --    & --    & --   & 63.20 \\
multiRAG & 52.84 & 66.12 & 53.63 & 70.15 & 67.19 & 77.52 & 68.89 & 67.32 & 52.21 & 67.96 & 66.46 & 59.44 & 66.57 & 70.86 \\
crossRAG & 54.63 & 66.12 & 55.31 & 70.15 & 69.38 & 78.50 & 68.13 & 69.61 & 46.18 & 66.77 & 64.18 & 62.24 & 66.57 & 68.03 \\
\ours{} (GPT-OSS-120B)  & 54.33 & 70.07 & 57.82 & 72.31 & 68.13 & 80.46 & 72.08 & 72.55 & 53.82 & 68.86 & 69.05 & 70.98 & 76.12 & 75.84 \\
\ours{} (Qwen3-235B) & 53.13 & 69.08 & 58.66 & 68.62 & 68.44 & 78.18 & 70.41 & 71.24 & 51.41 & 68.56 & 67.84 & 61.54 & 73.73 & 73.09 \\
\bottomrule
\end{tabular}
}
\caption{\textbf{Accuracy on cultural QA benchmarks with Ministral-3-8B-Instruct-2512 for a generator}. }
\label{tab:appendix_ministral-14B_table}
\end{table*}

\begin{table*}[t]
\resizebox{\textwidth}{!}{
\centering
\begin{tabular}{l ccccccccccccc c}
\toprule
\multirow{2}{*}{\textbf{Method}} &
\multicolumn{13}{c}{\textbf{BLEnD}} &
\multirow{2}{*}{\textbf{CLIcK}}\\
\cmidrule(lr){2-14}
& am & ar & as & az & el & en & es & fa & ha & id & ko & su & zh & \\

\midrule
Non-RAG  & 45.67 & 53.62 & 50.84 & 54.77 & 58.13 & 74.10 & 59.64 & 52.29 & 47.39 & 58.38 & 52.90 & 52.10 & 68.06 & 50.26 \\
monoRAG  & 51.34 & 61.51 & 48.88 & 58.77 & 63.44 & 68.89 & 64.04 & 60.78 & 50.60 & 63.17 & 60.67 & 55.94 & 64.78 & 58.88 \\
tRAG     & --    & --    & --    & --    & --    & --    & --    & --    & --    & --    & --    & --    & --   & 58.88 \\
multiRAG & 50.75 & 61.51 & 53.35 & 57.23 & 60.00 & 68.73 & 64.34 & 61.44 & 47.79 & 64.37 & 61.28 & 59.09 & 66.87 & 57.03 \\
crossRAG & 48.36 & 58.88 & 50.00 & 60.92 & 63.13 & 70.03 & 63.43 & 64.71 & 49.40 & 68.56 & 60.37 & 59.09 & 63.88 & 57.32 \\
\ours{} (GPT-OSS-120B)  & 49.55 & 64.14 & 59.78 & 64.62 & 65.31 & 74.76 & 67.37 & 66.67 & 53.01 & 66.17 & 67.38 & 68.53 & 71.94 & 62.08 \\
\ours{} (Qwen3-235B) & 51.64 & 64.14 & 54.75 & 61.23 & 64.69 & 76.22 & 67.37 & 65.69 & 59.44 & 68.56 & 66.31 & 63.99 & 73.13 & 61.86 \\
\bottomrule
\end{tabular}
}
\caption{\textbf{Accuracy on cultural QA benchmarks with Qwen3-1.7B for a generator}. }
\label{tab:appendix_qwen-1.7B_table}
\end{table*}

\begin{table*}[t]
\resizebox{\textwidth}{!}{
\centering
\begin{tabular}{l ccccccccccccc c}
\toprule
\multirow{2}{*}{\textbf{Method}} &
\multicolumn{13}{c}{\textbf{BLEnD}} &
\multirow{2}{*}{\textbf{CLIcK}}\\
\cmidrule(lr){2-14}
& am & ar & as & az & el & en & es & fa & ha & id & ko & su & zh & \\

\midrule
Non-RAG  & 55.82 & 63.82 & 59.50 & 67.69 & 65.63 & 82.41 & 71.02 & 66.99 & 53.01 & 67.07 & 65.85 & 62.59 & 77.91 & 58.96 \\
monoRAG  & 55.22 & 65.46 & 53.91 & 66.46 & 67.81 & 77.69 & 67.98 & 67.97 & 51.41 & 65.57 & 64.02 & 58.39 & 72.24 & 62.68 \\
tRAG     & --    & --    & --    & --    & --    & --    & --    & --    & --    & --    & --    & --    & --   & 62.68 \\
multiRAG & 55.52 & 66.45 & 54.19 & 67.08 & 69.38 & 75.73 & 71.02 & 70.59 & 52.21 & 69.16 & 63.72 & 63.29 & 72.24 & 70.11 \\
crossRAG & 55.52 & 66.12 & 56.70 & 65.54 & 66.56 & 78.18 & 69.35 & 71.24 & 53.01 & 68.86 & 64.18 & 64.34 & 72.54 & 69.07 \\
\ours{} (GPT-OSS-120B)  & 55.22 & 70.39 & 60.89 & 67.69 & 70.00 & 79.97 & 72.53 & 71.90 & 51.81 & 72.16 & 69.97 & 71.68 & 75.22 & 73.90 \\
\ours{} (Qwen3-235B) & 56.72 & 72.04 & 58.38 & 70.15 & 68.44 & 80.94 & 71.02 & 74.51 & 53.01 & 71.56 & 69.36 & 68.53 & 77.01 & 72.94 \\
\bottomrule
\end{tabular}
}
\caption{\textbf{Accuracy on cultural QA benchmarks with Qwen3-8B for a generator}. }
\label{tab:appendix_qwen-8B_table}
\end{table*}
\begin{table*}[t]
\resizebox{\textwidth}{!}{
\centering
\begin{tabular}{llccccccccccccc}
\toprule
\multirow{2}{*}{\textbf{Method}} &
\multirow{2}{*}{\textbf{Model}} &
\multicolumn{13}{c}{\textbf{BLEnD}} \\
\cmidrule(lr){3-15}
& & am & ar & as & az & el & en & es & fa & ha & id & ko & su & zh \\

\midrule
Self-RAG \cite{asai2024selfrag} & Llama-2-7B$^{\dagger}$ & 44.61 & 51.64 & 45.81 & 54.15 & 55.94 & 67.05 & 54.02 & 54.58 & 44.98 & 58.38 & 54.57 & 57.69 & 52.99 \\
\ours{} (Qwen3-235B) & Llama-2-7B-chat & 46.87 & 60.86 & 43.58 & 60.92 & 58.13 & 72.31 & 66.16 & 65.03 & 53.01 & 67.07 & 56.71 & 60.49 & 59.10 \\
\ours{} (Qwen3-235B) & Llama-3.2-3B & 55.22 & 66.45 & 60.03 & 68.31 & 69.69 & 79.32 & 71.32 & 72.55 & 58.23 & 71.26 & 68.60 & 64.69 & 74.04 \\
\bottomrule
\end{tabular}
}
\caption{\textbf{Comparison of agentic approaches on BLEnD}. For all methods, the maximum token budget was set to 1024. To comply with the model’s context length, the number of retrieved documents (top-k) was limited to 1. $^{\dagger}$ indicates that the model was trained by the authors separately. Queries exceeding the model's context length were excluded from evaluation.}
\label{tab:appendix_SelfRAG_comparison_table}
\end{table*}

\begin{figure*}[t]
\centering
\begin{minipage}{\textwidth}
\begin{mdframed}

SYSTEM PROMPT:\\
You are a helpful AI Assistant with expertise in cultural and linguistic content classification, acting as the **search orchestrator** of a multi-corpus Retrieval-Augmented Generation (RAG) system.\\
\\
\texttt{[Your Task]}\\
Given an input query (which may include a passage, a question, and optionally, multiple-choice options), you must **Select language corpora** to search.  \\
\\
\texttt{[Corpus selection rules]}\\
1. Always include the corpus whose language code matches the primary language of the query.\\
2. If some corpora are **content-wise** relevant (country, region, culture, institution, person, etc.), you may additionally select them.\\
\\
   - The query explicitly contains terms in another language or the user's intent clearly benefits from cross-language retrieval (e.g., looking for translations, comparative cultural information).\\
   - Example: A topic about Japan → select "ja".\\
\\
3. Do not select corpora that are almost unrelated to the query.\\
4. **Never** add a corpus "just in case". Choose only a small, realistically useful set.\\
5. Use only language codes that appear in the following langauge pools. **Never invent new names**.\\
Language Pools: $[$"id", "am", "su", "ar", "ha", "en", "zh", "ko", "as", "el", "fa", "es", "az"$]$\\
\\
\texttt{[Output format]}\\
Return **exactly** the following JSON object **as a single continuous line with no surrounding whitespace, line breaks, or markdown formatting**:\\
\\
\{"language\_names": $[$"<lang\_code>", ... $]$\}
\begin{itemize}
    \item  \texttt{language\_names} must be a list of **valid** language codes from the pool, containing **at most three** entries and **always** including the primary language of the query.
\end{itemize}
USER PROMPT:\\
$[$USER QUERY$]$
\{USER\_QUERY\}
\end{mdframed}
\end{minipage}
\caption{\textbf{Planner Prompt template.}}
\label{fig:planner-prompt-template}
\end{figure*}


\begin{figure*}[t]
\centering
\begin{minipage}{\textwidth}
\begin{mdframed}
SYSTEM PROMPT:\\
You are a helpful AI Assistant with expertise in cultural and linguistic content classification, acting as the "second-stage search orchestrator" of a multi-corpus Retrieval-Augmented Generation (RAG) system.\\
\texttt{[Your Task]}\\
You are given:
\\
   - the original input query (which may include a passage, a question, and optionally, multiple-choice options),\\
   - the previously used rewritten query for retrieval,\\
   - the previously chosen language codes for retrieval,\\
   - the system's reasoning explaining why the former retrieval attempt was not sufficient.\\
\\
Your job is to:\\
1. **Select language corpora** for the next retrieval round.\\
2. **Rewrite the query** to improve retrieval quality, grounded in the system's reasoning.\\
\\
You MUST NOT simply repeat the previous decision.  \\
At least one of the following must change:\\
\\
   - the set of language codes (`language\_names`), OR\\
   - the rewritten query (focus, structure, or keywords).\\
\\
\texttt{[Corpus selection rules]}\\
1. Always include the corpus whose language code matches the **primary language** of the original query, unless the system's reasoning explicitly shows it is consistently low-relevance.\\
2. If some corpora are **content-wise** relevant (country, region, culture, institution, person, event, etc.), you may additionally select them.
\\
   - Example: a topic about Japan → include "ja".\\
\\
3. If the system's reasoning indicates that many documents from a language were off-topic, shallow, or irrelevant, you may lower its priority or remove it, and instead consider other content-relevant languages.\\
4. Do not select corpora that are almost unrelated to the query.\\
5. **Never** add corpora "just in case." Choose only a small, realistically useful set.\\
6. Use only language codes that appear in the following language pools. **Never invent new names.**\\
   Language Pools: $[$"id", "am", "su", "ar", "ha", "en", "zh", "ko", "as", "el", "fa", "es", "az"$]$\\
\\
\end{mdframed}
\end{minipage}
\caption{\textbf{Planner Prompt Template w/ critique.}}
\label{fig:planner-prompt-template-w-critique}
\end{figure*}

\begin{figure*}[t]\ContinuedFloat
\centering
\begin{minipage}{\textwidth}
\begin{mdframed}
\texttt{[Query rewriting rules]}\\
1. **Preserve the original meaning and intent**, while making the query clearer and more retrieval-friendly:\\
   - Remove colloquial or filler phrases.\\
   - Explicitly mention time, location, and named entities ONLY when given. Do not add unnecessary details.\\
   - **Do not delete any complete sentences in the original query that convey substantive information** (given passage, main question, etc.).\\
   - Remember that the rewritten query is the only source of information for the retriever.\\
\\
2. Adjust the rewritten query using the system's reasoning:\\
   - If results were too broad → make the query more specific.\\
   - If important aspects were missing → add them explicitly.\\
   - If results were off-topic → clarify the main topic and disambiguate the concepts.\\
   - If the structure was unclear → reorganize for better retrieval.\\
3. The new rewritten query must **meaningfully differ** from the previous rewritten query (e.g., emphasize a different aspect, add missing constraints, reorganize structure, clarify ambiguous elements).\\
\\
\texttt{[Output format]}\\
Return **exactly** the following JSON object **as a single continuous line with no surrounding whitespace, line breaks, or markdown formatting**:\\
\\
\{\\
  "language\_names": $[$"<lang\_code>", ... $]$,\\
  "rewritten\_query": "<cleaned, rewritten query>"\\
\}\\
\\
- `language\_names` must be a list of **valid** language codes from the pool, containing **at most three** entries and always including the primary language of the original query unless the system's reasoning indicates otherwise.\\
- `rewritten\_query` must be a single string (may be empty).
\\
\\
USER PROMPT:\\
\texttt{[ORIGINAL USER QUERY]}\\
\{USER\_QUERY\}\\
\texttt{[PREVIOUS QUERY FOR RETRIEVAL]}\\
\{REWRITTEN\_QUERY\}\\
\texttt{[PREVIOUS LANGUAGE CORPORA FOR RETRIEVAL]}\\
\{PREV\_LANGS\}\\
\texttt{[REASON FOR ADDITIONAL RETRIEVAL]}\\
\{REASEON\}
\end{mdframed}
\end{minipage}
\caption{\textbf{Planner Prompt template w/ critique. (continued)}}
\label{fig:planner-prompt-template-w-critique-continued}
\end{figure*}


\begin{figure*}[t]
\centering
\begin{minipage}{\textwidth}
\begin{mdframed}
SYSTEM PROMPT:\\
You are a document re-ranking system. Your role is to evaluate a user query and a set of retrieved candidate documents. For each document, you must infer several properties, assign numerical scores based on the rubric, and provide a final evaluation. Your evaluation focuses on how well each document contributes to answering the user's query—especially in multilingual or cross-domain scenarios.\\
\\
\texttt{[Inferred Properties]}\\
Relevance (0-5)\\
- Measures how strongly the document aligns with the key concepts, entities, and intent of the query.\\
- Higher scores correspond to closer conceptual alignment and direct topical relevance.\\
- Lower scores correspond to weak or minimal connection to the query.\\
\\
Usefulness (0-5)\\
- Measures how much the document helps the system construct a correct, complete, and actionable answer.\\
- Higher scores indicate substantial, high-impact contributions.\\
- Lower scores indicate little to no helpful information.\\
\\
Clarity and Specificity (0-5)\\
- Measures how clearly, precisely, and unambiguously the document presents information relevant to the query.\\
- Higher scores reflect well-structured, specific, and easy-to-interpret content.\\
- Lower scores reflect vague, overly general, or confusing content.\\
\\
Compatibility (0-5)\\
- Measures linguistic, cultural, and domain compatibility between the query and the document.\\
- Higher scores correspond to strong language alignment, faithful cross-lingual equivalence, and contextual appropriateness.\\
- Lower scores correspond to mismatched languages, cultural contexts, or domain assumptions.\\
\\
\texttt{[Output Format]}\\
You must output **ONLY ONE** JSON dictionary corresponding to the evaluation of a **single document**, with **no additional text, no explanations, no Markdown, and no commentary**.\\
The JSON must follow **exactly** this structure:\\
\\
\{"scores": \{"relevance": RELEVANCE\_SCORE(0-5), "usefulness": USEFULNESS\_SCORE(0-5), "clarity\_specificity": CLARITY\_SPECIFICITY\_SCORE(0-5), "compatibility": COMPATIBILITY\_SCORE(0-5)\}, "critique": "CRITIQUE\_TEXT"\}\\
\\
Strict requirements:\\
- All scores must be integers from 0 to 5.\\
- "critique" must be based on the content of the given document without any hallucinations and be a single string describing the reasoning.\\
- **No other hierarchies, nested structures, arrays, multiple document keys, or additional fields are allowed.**\\
- Do NOT wrap the output in other objects.\\
- Do NOT output multiple dictionaries.\\
- Do NOT include the document ID, name, or any other label as a key.\\
- Do NOT output anything before or after the JSON dictionary.
\end{mdframed}
\end{minipage}
\caption{\textbf{Critique Prompt Template for Scoring.}}
\label{fig:critique-prompt-template-scoring}
\end{figure*}


\begin{figure*}[t]
\centering
\begin{minipage}{\textwidth}
\begin{mdframed}
SYSTEM PROMPT:\\
You are a retrieval controller for a RAG system.\\
\texttt{[Your job]}\\
Given a user query and a set of retrieved documents, decide whether these documents are sufficient to answer the query reliably, and which documents are actually useful.\\
\\
\#\# Inputs\\
\texttt{[Query]}\\
- content: text\\
\\
\texttt{[Retrieved Documents]}\\
Each document has: \\
- content: text\\
- scores: a numeric score (higher means more relevant)\\
- critique: natural language explanation of why this document may be appropriate or sufficient for answering the query.\\
\#\# Decision Guidelines\\
Only consider the information available in the documents, and do not use external knowledge.
When making your decision, consider:\\
\\
1. Coverage\\
   - Do the given documents collectively cover the main aspects and requirements of the query?\\
   - Are there important sub-questions or constraints in the query that are not addressed?\\
   - Are all information and details of the documents considered to solve the problem?\\
\\
2. Depth \& Specificity\\
   - Are the documents detailed and specific enough to support a precise and reliable answer?\\
   - If the query requires factual accuracy, step-by-step reasoning, or up-to-date information, be conservative: if you are not confident, prefer requesting more documents.\\
\\
3. Consistency\\
   - Do the documents agree on key facts?\\
   - If there are major contradictions that you cannot resolve with the current documents, you may need more documents.\\
\\
\#\# Output Format\\
Respond in **valid JSON** with the following fields:
- "enough\_documents": boolean\\
    - true  = the given documents are sufficient to answer the query reliably  \\
    - false = you believe we should retrieve more documents\\
- "reason": text\\
    - reason why you think the documents are sufficient or not. If they are insufficient, additiional feedback for next search will be needed.\\
\{
  "enough\_documents": true/false,
  "reason": "reason for your decision"
\}\\
\\
USER PROMPT:\\
\texttt{[USER QUERY]}\\
\{QUERY\}\\
\texttt{[RETRIEVED DOCUMENTS]}\\
\{DOCUMENTS\}\\

\end{mdframed}
\end{minipage}
\caption{\textbf{Critique Prompt Template for evaluating sufficiency.}}
\label{fig:critique-prompt-template-sufficiency}
\end{figure*}


\begin{figure*}[t]
\centering
\begin{minipage}{\textwidth}
\begin{mdframed}
\textbf{Original Query}:\\
다음은 한국의 문화와 전통에 대한 문제이다.\\
제사에서 가족이 하는 행동은 무엇인가?\\
A: 조상에게 절을 두 번 한다\\
B: 조상에게 절을 세 번 한다\\
C: 조상에게 절을 한 번 한다\\
D: 조상에게 절을 네 번 한다\\
\textcolor{gray}{\small The following is a question about Korean culture and traditions. What do family members do during Jesa}\\
\textcolor{gray}{\small A: Bow twice to the ancestors}\\
\textcolor{gray}{\small B: Bow three times to the ancestors}\\
\textcolor{gray}{\small C: Bow once times to the ancestors}\\
\textcolor{gray}{\small D: Bow four times to the ancestors}\\

\textbf{Answer}: A\\

\hrule
\vspace{0.8em}

\textbf{Retrieved Documents from $\mathcal{C}_{all}$}

\textbf{Document 1:} 
한국의 제사(韓國의 祭祀)는 크게 천지신명을 비롯한 자연물에 드리는 제사와 조상에게 드리는 제사로 나뉜다. ...\\
\textcolor{gray}{\small Korean \textit{Jesa} (ancestral rites) is broadly categorized into rites performed for nature, including the gods of heaven and earth, and rites performed for ancestors. ...}\par\vspace*{0.3em}

\textbf{Document 2:} 
La première partie de la célébration du dol est la prière. ...\\
\textcolor{gray}{\small The first part of the \textit{Dol} celebration is prayer. ...}\par\vspace*{0.3em}

\textbf{Document 3:} 
... \#\# 한국의 제사 ... \#\# 힌두교의 제사 ...\\
\textcolor{gray}{\small ... \#\# Korean \textit{Jesa} ... \#\# Hindu rites ...}\par\vspace*{0.3em}

\textbf{Document 4:} 
\#\#\# Social folk customs in daily life ...\par\vspace*{0.3em}

\textbf{Document 5:} 
... Charye is one of the ancestral memorial rites celebrated during Chuseok, ...

\vspace{0.8em}
\hrule
\vspace{0.8em}

\textbf{Retrieved Documents form \ours{}}

  \textbf{Document 1:} 
  ... 제주가 \textbf{두 번 절한다.} ...\\
  \textcolor{gray}{\small ... The chief mourner bows twice. ...}\\
$\cdots$

\end{mdframed}
\end{minipage}
\caption{\textbf{Planner Critique Example for Retrieved Documents.} Qualitative comparison of retrieved evidence for a culturally grounded Korean query (jesa bowing practice). Retrieval over the unified corpus $\mathcal{C}_{all}$ produces mostly superficial or tangential documents, reflecting substantial noise from indiscriminate corpus expansion. In contrast, \ours{} routes retrieval to a linguistically and culturally aligned Korean document that explicitly describes the jesa procedure, including the correct two-bow ritual, enabling access to precise procedural knowledge needed to answer correctly.}
\label{fig:corpus-noisy-example}
\end{figure*}

\begin{figure*}[t]
\centering
\begin{minipage}{\textwidth}
\begin{mdframed}
\textbf{Original Query}:\\
What region in the US is usually associated with oil?\\

\textbf{Documents (1st Trial)}

\textbf{Document 1:} Hassi Messaoud () is a town in Ouargla Province, eastern Algeria, locatedsoutheast of Ouargla. As of 2008 it had a population of 45,147 people, ...\par\vspace*{0.3em}

\textbf{Document 2:} The above north-south Algerian road from Constantine passes through other oases. North of Wargla [Ouargla] lies Touggourt [Tuggurt, Taghit] (pop: 153,000), ...\par\vspace*{0.3em}

\textbf{Document 3:} Andalus may refer to: \#\# PlacesAl-Andalus, a historical region in Europe around the Iberian PeninsulaAndalusia, ...\par\vspace*{0.3em}

\textbf{Document 4:}\#\# Economy The economy of Patos is mainly based on oil companies such as Bankers Petroleum, and Albpetrol. Patos is on the Patos-Marinza Oil Field ...\par\vspace*{0.3em}

\textbf{Document 5:}Hassi Messaoud Oil Field is an oil field located in Ouargla Province. It was discovered in 1956 by S.N. REPAL and developed by Sonatrach. ...\par\vspace*{0.6em}

\textbf{Critique Decision}\\
 \textit{enough\_documents:} False\\
 \textit{reason:} No documents were retrieved, so there is no information available to determine which region in the US is associated with oil. Additional documents are needed to answer the question reliably.\\

\vspace{0.4em}
\hrule
\vspace{0.4em}

\textbf{Rewritten Query}:
Which region in the United States is most commonly associated with oil production or the oil industry?\\

\textbf{Documents (2nd Trial)}

\textbf{Document 1:} ... The leading crude oil-producing areas in the United States in 2023 were Texas, followed by the offshore federal zone of the Gulf of Mexico, North Dakota and New Mexico. ...\par\vspace*{0.3em}

\textbf{Document 2:} ... The Glenn Pool strike near Tulsa, Oklahoma in 1905 established Tulsa as the leading U.S. oil production center until the 1930s. Though Texas soon lagged behind Oklahoma and California, it was still a major producer. ...\par\vspace*{0.3em}

\textbf{Document 3:} Oil is a major part of the Texas economy. Texas has consistently been the leading oil producing state ...\par\vspace*{0.3em}

\textbf{Document 4:} ... La principal zona productora de petróleo de Estados Unidos en 2019 fue Texas ...\\
\textcolor{gray}{\small ... The main oil-producing region in the United States in 2019 was Texas ...}\par\vspace*{0.3em}

\textbf{Document 5:} ... 自从在纺锤顶发现石油以来，能源一直是德克萨斯政治和经济上的主导力量，如果将德州视为国家，它将成为世界第六大石油生产国。德州已知的石油储量约为，约占-{}-美国已知储量的四分之一。...\\
\textcolor{gray}{\small ... Since the discovery of oil at Spindletop, energy has been a dominant force in Texas's politics and economy. If Texas were considered a nation, it would be the world's sixth-largest oil producer. Texas's known oil reserves are approximately [amount missing], representing about a quarter of the total known oil reserves in the United States. ...}\par\vspace*{0.6em}

\textbf{Critique Decision}\\
 \textit{enough\_documents:} True\\
 \textit{reason:} The retrieved documents collectively provide strong and consistent evidence that the southwestern region of the United States—particularly Texas, New Mexico, and the Gulf Coast—is most commonly associated with oil production. ... Therefore, no further documents are needed.\\

\end{mdframed}
\end{minipage}
\caption{\textbf{Planner Critique Example for Query Rewriting and Evidence Refinement.} The critic identifies insufficient information in the initial retrieval and rejects the evidence. Following a planner-led query rewrite to include missing signals, the second retrieval provides specific details on the U.S. oil industry, enabling an accurate response.}
\label{fig:query-rewrite-example}
\end{figure*}


\section{Use Of AI Assistants}
The authors used AI assistants for the language of the paper and codes for the experiments.

\end{document}